\documentclass[runningheads]{llncs}

\usepackage{eccv}

\usepackage{eccvabbrv}

\usepackage{graphicx}
\usepackage{booktabs}

\usepackage[accsupp]{axessibility}  %

\usepackage{hyperref}

\usepackage{orcidlink}

\usepackage{float}

\begin{document}

\title{GeneralAD: Anomaly Detection Across Domains by Attending to Distorted Features} 

\titlerunning{GeneralAD}

\author{Luc P.J. Sträter\textsuperscript{*} \orcidlink{0009-0006-2631-4499} \and
Mohammadreza Salehi\textsuperscript{*} \orcidlink{0000-0002-9247-9439} \and\\
Efstratios Gavves \orcidlink{0000-0001-8947-1332} \and 
Cees G. M. Snoek \orcidlink{0000-0001-9092-1556} \and
Yuki M. Asano \orcidlink{0000-0002-8533-4020}
}

\authorrunning{L.P.J.~Sträter et al.}

\institute{University of Amsterdam, The Netherlands \\
\email{lucstrater@gmail.com}, \email{s.salehidehnavi@uva.nl}}

\maketitle

\begin{abstract}
In the domain of anomaly detection, methods often excel in either high-level semantic or low-level industrial benchmarks, rarely achieving cross-domain proficiency. Semantic anomalies are novelties that differ in meaning from the training set, like unseen objects in self-driving cars. In contrast, industrial anomalies are subtle defects that preserve semantic meaning, such as cracks in airplane components. In this paper, we present GeneralAD, an anomaly detection framework designed to operate in semantic, near-distribution, and industrial settings with minimal per-task adjustments. In our approach, we capitalize on the inherent design of Vision Transformers, which are trained on image patches, thereby ensuring that the last hidden states retain a patch-based structure. We propose a novel self-supervised anomaly generation module that employs straightforward operations like noise addition and shuffling to patch features to construct pseudo-abnormal samples. These features are fed to an attention-based discriminator, which is trained to score every patch in the image. With this, our method can both accurately identify anomalies at the image level and also generate interpretable anomaly maps. 
We extensively evaluated our approach on ten datasets, achieving state-of-the-art results in six and on-par performance in the remaining for both localization and detection tasks. Code available at \url{https://github.com/LucStrater/GeneralAD}.

\end{abstract}

\def\thefootnote{\fnsymbol{footnote}}
\footnotetext[1]{These authors contributed equally to this work.}
\def\thefootnote{\arabic{footnote}}

\section{Introduction}
\label{sec:intro}
In Anomaly Detection (AD), the task is to learn the distribution of a given training dataset and distinguish any test sample that does not belong to it. 
Sample applications vary widely, including the detection of anomalous objects in self-driving cars, which typically operate at a semantic level, and the identification of defects in industrial assembly lines, where the focus is more on low-level elements like pixels~\cite{salehi2021unified}. During training, all the samples are labeled as ``normal'', and there is no access to ``abnormal'' inputs, which can be formulated similarly to one-class learning tasks. 
Given this unusual setting of having no access to a complete category that is relevant for testing, this field has spawned many approaches.
For example,~\cite{mirzaei2023fake, cohen2022transformaly, salehi2021multiresolution, deng2022anomaly, reiss2023mean} have employed the benefits of pretrained features and diffusion models to learn normal representations that are capable of better solving semantic tasks. 
Orthogonal to this,~\cite{liu2023simplenet, gu2023remembering, bae2023pni, zhang2023destseg, zhang2023unsupervised} have proposed specific architectures to tackle industrial benchmarks for both detection and localization of anomalies.

However, despite the encouraging progress toward better modeling of normal distributions, a growing chasm has been created between the methods that perform well on semantic benchmarks vs on industrial ones. For instance, leading models in semantic benchmarks, such as Transformaly~\cite{cohen2022transformaly} and MSAD~\cite{reiss2023mean}, demonstrate a significant drop in performance when applied to industrial benchmarks like MVTec-AD~\cite{mvtecad}. This trend is mirrored in the context of industrial-focused models like SimpleNet~\cite{liu2023simplenet} and Recontrast~\cite{guo2024recontrast}, which show similar underperformance on semantic datasets, including CIFAR-10~\cite{cifar}. Despite both sets of methods often utilizing the same strong pretrained features, their one-sided metiers areas suggest overfitting of methodology for specific datasets. 

In this work, we aim to produce an all-rounder model that performs well across different tasks and datasets with minimum per-task modification, towards General Anomaly Detection (GeneralAD). Starting with a pretrained Vision Transformer~\cite{oquab2023dinov2} feature extractor, first, we introduce a self-supervised anomaly feature generation module, which gets features of normal samples as the input and generates abnormal ones by applying simple operations such as adding noise and shuffling patches. This results in the generation of high-quality pseudo-abnormal samples, which are not easily identifiable due to the selection of the small noise magnitudes. Moreover, the shuffling introduces logical anomalies, further complicating their detection. Second, we propose to use a transformer-based discriminator, which takes patch features as the input and is trained to detect structural and logical anomalies by attending to different features at different locations. This creates a versatile anomaly detection discriminator, capable of identifying anomalies at varying levels as required, ranging from patch-level to image-level anomalies. Finally, our method is not only able to detect anomalies at the image level but also produce interpretable anomaly maps that can be used to pinpoint abnormal patches for both semantic and industrial tasks.

We evaluate the model across ten different datasets from different benchmarks such as CIFAR-10~\cite{cifar}, CIFAR-100~\cite{cifar}, Fashion-MNIST~\cite{fmnist} and View~\cite{view} for semantic anomaly detection, Aircraft-FGVC~\cite{fgvcaircraft} and Stanford-Cars~\cite{stanfordcars} for near anomaly detection~\cite{mirzaei2023fake}, where anomaly distribution is very close to normal one, and MVTec-AD~\cite{mvtecad}, MVTec-LOCO~\cite{mvtecloco}, VisA~\cite{visa}, and MPDD~\cite{mpdd} for industrial anomaly detection. Our results show that the proposed method matches state-of-the-art performance on datasets like FMNIST~\cite{fmnist}, Stanford-Cars~\cite{stanfordcars}, MVTec-AD~\cite{mvtecad}, and VisA~\cite{visa}, and surpasses it in both detection and localization on all other datasets. Overall, this paper makes the following contributions:

\begin{itemize}
    \item We introduce a self-supervised anomaly feature generation module that mimics a wide range of anomalies, from pixel-level to semantic, by applying a simple and diverse set of operations in the feature space.
    \item We propose a transformer-based discriminator that effectively operates from individual patches to subsets of patches and entire images. This enables the discriminator to identify and pinpoint a wide range of anomalies in the input data, including structural, logical, and semantic inconsistencies.
    \item We evaluate our method by conducting comprehensive experiments on three different benchmarks and ten datasets, achieving state-of-the-art results in six out of ten cases. This shows the generality and applicability of the model for different tasks with a minimum amount of per-task modification.
\end{itemize}

\section{Related Work}
\label{sec:related_work}
Anomaly detection encompasses various types of irregularities, each traditionally requiring specific detection methods. In this section we show the distinct areas that intersect in our work and discuss related methods. 

\subsection{Semantic Anomaly Detection.}
Methods in the semantic setting are designed to identify novelties that deviate semantically from the training data distribution. When the deviation between the normal and abnormal distributions in the evaluation dataset is large, it is referred to as an anomaly detection task, as seen in the CIFAR-10 dataset. Conversely, if the deviation is small, it is termed near anomaly detection, exemplified by the Aircraft-FGVC dataset~\cite{mirzaei2023fake}. Common solutions in this domain include self-supervised learning methods~\cite{golan2018deep,hendrycks2019using,reiss2023mean,tack2020csi} and leveraging features from pretrained models~\cite{salehi2021multiresolution, mirzaei2023fake,reiss2021panda,cohen2022transformaly}. Transformaly~\cite{cohen2022transformaly}, for instance, extends the student-teacher architecture proposed in KDAD~\cite{salehi2021multiresolution} and achieves state-of-the-art results on semantic anomaly detection benchmarks. FYTIMI~\cite{mirzaei2023fake}, the state-of-the-art in near semantic anomaly detection, uses diffusion models~\cite{ho2020denoising} to generate pseudo-anomalies, which are then used to solve a binary classification task between the normal dataset and the generated abnormal inputs. Although these methods have achieved superior results on semantic datasets, their ability to detect patch-level anomalies (defections) is limited, resulting in poor performance on industrial benchmarks. 

\subsection{Industrial Anomaly Detection.} 
Methods in the industrial setting use intact training samples to identify defects during testing. It typically focuses on fine-grained anomalies, where only small parts of an image differ from the norm. This distinguishes it from semantic and near anomaly detection, where anomalies span the entire image. There are two main approaches: synthesizing-based~\cite{li2021cutpaste,zavrtanik2021draem} and embedding-based~\cite{guo2024recontrast, roth2022towards, xie2022rdad, gudovskiy2022cflow, defard2021padim, reiss2021panda, batzner2024efficientad, cao2023anomaly,cohen2020sub}, methods. Synthesizing-based methods attempt to approximate the abnormal distribution by employing augmentations~\cite{li2021cutpaste} or feature space manipulation to facilitate the detection task~\cite{liu2023simplenet}.  Embedding-based methods, such as Recontrast~\cite{guo2024recontrast} and PatchCore~\cite{roth2022towards}, focus on transforming normal image features into a space where anomalies stand out. While these approaches excel at identifying pixel-level anomalies, they are less effective at detecting semantic ones. For instance, SimpleNet~\cite{liu2023simplenet}, the state-of-the-art model for this benchmark, is designed to detect anomalies only at the level of local patches. Consequently, it lacks a global perspective necessary for modeling cross-correlations between patches, which is essential for effective semantic anomaly detection. GeneralAD, however, is designed to detect anomalies not only at the patch level but also by identifying abnormal correlations between patch features. This capability makes it a versatile solution for various types of anomalies.

\subsection{Self-supervised Anomaly Detection}
Self-supervised learning approaches have been applied to detect both semantic anomalies \cite{golan2018deep,hendrycks2019using,reiss2023mean,tack2020csi, cai2022perturbation} and industrial anomalies \cite{li2021cutpaste,liu2023simplenet}. However, the way the proxy task is set up differs depending on the type of anomaly being targeted. For semantic anomalies, the proxy task often involves learning high-level features that capture the normal data distribution, such as through rotation prediction \cite{golan2018deep}, perturbation prediction \cite{cai2022perturbation} or contrastive learning \cite{tack2020csi}. On the other hand, for industrial anomalies, the proxy task typically focuses on learning low-level features that are sensitive to local irregularities, such as through patch-based masking \cite{li2021cutpaste}. GeneralAD is designed to predict whether pretrained features have been edited, with the set of editing operations tailored to the type of anomaly to be detected.

\section{Method}
\label{sec:method}
The problem of {\it Anomaly Detection} is defined using the following setup \cite{chandola2009anomaly, chalapathy2019deep, pang2021deep, salehi2021unified, liu2024deep}. Let $\mathcal{P}$ be a joint probability distribution defined over the input space $\mathcal{X}$ and the output space $\mathcal{Y}$. The output space $\mathcal{Y}$ contains only one class label, which is ``normal'', i.e., $|\mathcal{Y}| = 1$. Let $\mathcal{D}_{\text{Normal}}$ denote the marginal distribution of $\mathcal{P}$ over $\mathcal{X}$. A neural network $\mathcal{F}$ is trained on samples drawn from $\mathcal{D}_{\text{Normal}}$ to generate a feature embedding, which is subsequently used to compute the anomaly score of an input sample. The objective of anomaly detection is to construct a decision function $\mathcal{M}$ such that for any given test input $x \in \mathcal{X}$ \autoref{eq:decision_function} holds.

\begin{equation}
\label{eq:decision_function}
\mathcal{M}(x, \mathcal{F})=
\begin{cases}
0 & \text{if}\ x \sim \mathcal{D}_{\text{Normal}}\text{,} \\
1 & \text{otherwise.}
\end{cases}
\end{equation}

\begin{figure}
  \centering
  \includegraphics[width=0.99\textwidth]{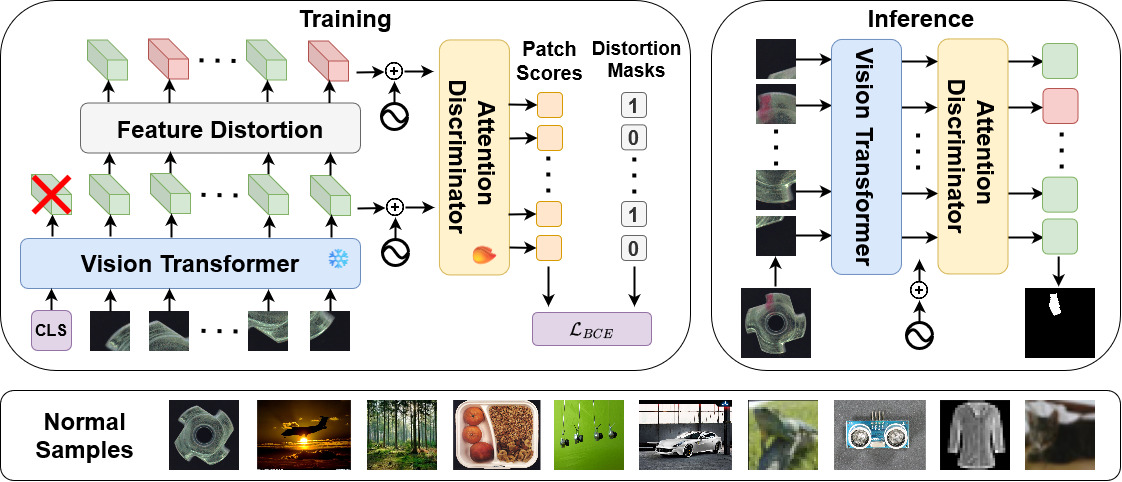}
  \caption{\textbf{The proposed method overview.} During the training, first, an image is segmented into smaller patches and passed to the pretrained vision encoder to extract patch features. Next, the extracted patches are distorted by the self-supervised anomaly feature generation module, labeled as feature distortion. Finally, all the patches are passed to the cross-patch attention discriminator, whose role is to detect semantic, logical, or structural distortions. 
  During the inference, the feature distortion module is deactivated, and the discriminator is used for both detection and localization tasks. The lower part of the figure displays various normal distributions.}
  \label{fig:method}
\end{figure}

Our method aims to make a framework in which, the decision function~$\mathcal{M}$ is trained without relying on specific designs for particular datasets or tasks. To do so, we assume current state-of-the-art pretrained networks~\cite{caron2021emerging, oquab2023dinov2} can be used to extract semantic feature representations as done by $\mathcal{F}$. Now, to train $\mathcal{M}$, we need access to both normal and abnormal samples, which is not the case in our setup. To solve the issue, we attempt to approximate them by proposing the self-supervised anomaly feature generation module. This module creates abnormal features by making structural or logical anomalies in the feature space.  Finally, to find a decision function~$\mathcal{M}$ with minimal design biases, positional embeddings are added to the features, and they are passed to a cross-patch attention discriminator module, which employs a multi-head attention mechanism to detect abnormal regions. During inference time, each input is first passed to the feature extractor, then the extracted patch features are forwarded to the discriminator to get per-patch anomaly scores. For image-level scores, the average of a subset of the patches is considered as the abnormality score. Our method is visualized in Figure~\ref{fig:method} and explained in more detail in the following parts.

\subsection{Feature Extraction}
Consider an image represented as $x \in \mathbb{R}^{3 \times H \times W}$, which is  decomposed into patches of size $P \times P$, resulting in $N {=} \frac{H}{P} {\times} \frac{W}{P}$ patches, labeled as $x^{p}_j, j \in {1, \ldots, N}$. These patches are then processed through a pre-trained encoder $\mathcal{F}$, generating a set of spatial tokens and one classification token. As the classification token is specifically tailored for semantic tasks, and our method aims to perform well across various tasks, we will exclusively utilize spatial tokens expressed as $\mathcal{F}(x) {=} [{f_1, \ldots, f_N}]$. In addition to spatial tokens, we extract their corresponding attention maps for each head $i$, $\mathcal{A}_i(x) = [a_1, \ldots, a_N]$, to better utilize more informative regions. Here, the attention value for each patch is with respect to the classification token. For subsequent stages in our pipeline, the goal is to learn a decision function $\mathcal{M}$ defined over $\mathcal{F}(x)$, which generalizes well to unseen normal test time inputs yet can not do the same for abnormal ones.

\subsection{Self-supervised Anomaly Feature Generation}
Having extracted pretrained features $\mathcal{F}(x)$ and the attention maps $\mathcal{A}_i(x)$, we know that patch representations are semantic~\cite{caron2021emerging, salehi2023time, ziegler2022self}; therefore, they can generalize to unseen normal inputs well. Yet, refining generalization boundaries often necessitates access to abnormal distributions, a step unfeasible in our context. Existing studies, such as~\cite{mirzaei2023fake}, try to create these distributions through the outputs of early-stage trained diffusion models. However, this approach is costly and fails to address structural or logical anomalies usually found in industrial applications. To address this limitation, we propose a self-supervised anomaly feature generation module ($\text{SAG}$), which receives $\mathcal{F}(x)$ and $\mathcal{A}_i(x)$ as the input and generates anomalies in the feature space by adding noise to random locations or copy-pasting features to strongly attended regions. The resulting \textit{distorted} features are called $\mathcal{F}\epsilon(x)$, and the map of distorted locations is called the distortion mask. The distortion parameters are adjusted by $\epsilon$ in $\mathcal{F}\epsilon(x)$:

\begin{equation}
    \mathcal{F}\epsilon(x) {=}  \text{SAG}(\mathcal{F}(x), \mathcal{A}_i(x)).
\end{equation}

We explore three distinct distortion strategies to obtain $\mathcal{F}\epsilon(x)$: (1) \textit{Noise All Patches}, (2) \textit{Noise Random Patches}, and (3) \textit{Attention Shuffle}. The first strategy involves adding Gaussian noise to the features of all patches in the input, while the second strategy applies Gaussian noise to a randomly selected fraction of patches. In the third strategy, important patches are selected by uniformly sampling an attention head from the backbone and identifying $N$ patches with the highest attention values. Here, the attention value for each patch is taken with respect to the classification token and $N$ is uniformly sampled between 1 and the total number of patches. These patches are then shuffled in the feature space.

\subsection{Cross-Patch Attention Discriminator}
Upon generating anomalous feature maps, denoted as $\mathcal{F}_\epsilon(x)$, and their normal counterparts $\mathcal{F}(x)$, our objective is to effectively train a discriminator. This discriminator, when presented with an input $I$—which could be either $\mathcal{F}_\epsilon(x)$ or $\mathcal{F}(x)$—should be capable of identifying anomalous patches. In pursuit of ensuring robust performance across a diverse range of anomaly types, it is imperative for the discriminator to discern semantic, structural, or logical irregularities present in $\mathcal{F}_\epsilon(x)$. To achieve this, we propose an approach that initially employs a Multi-Head Attention (MHA) mechanism. This mechanism is designed to combine information from all patch features. Subsequently, these fused patch features are processed through a Multi-Layer Perceptron (MLP) network, which is responsible for calculating the anomaly score for each individual patch $i$ shown by $P_{\text{score}}(i)$. Finally, positional embeddings $E_{\text{pos}}$ are added to the discriminator's input to give it a sense of the input feature positions:
\begin{align}
     F_{\text{fused}}&{=} \text{MHA}(I+E_{\text{pos}}) ,\\
    P_{\text{score}}&{=} \text{MLP}(\text{LN}(F_{\text{fused}})).
\end{align}
The training objective function is defined as the cross entropy loss between the patch scores and distortion masks.

\subsection{Inference}
During testing, we derive an image-level anomaly score from patch-level scores by picking the top $K$ highest patch scores and averaging them. This choice of $K$ depends on the size of the anomaly region in the dataset. 

\begin{equation}
    I_{\text{score}} =  \frac{1}{K} \sum_{i=1}^{\text{Top }K} P_{\text{score}}(i) .
\end{equation}

\section{Experiments}
\label{sec:experiments}
In the subsequent sub-sections, we present the datasets, implementation details, results, and ablation study. 

\subsection{Datasets}
Our experiments encompass three distinct benchmarks. First, we evaluate our approach on four \textit{semantic anomaly detection} datasets: CIFAR-10~\cite{cifar}, CIFAR-100~\cite{cifar}, Fashion-MNIST~\cite{fmnist}, and View~\cite{view}. In this setting, the model is trained using a single normal class from the dataset and subsequently tested against the remaining classes, which are treated as anomalies. Anomalies in these datasets typically span the entire image.

Second, we explore \textit{near anomaly detection}, focusing on identifying subtle deviations within datasets \cite{mirzaei2023fake}. We test on the two main benchmarks proposed for this purpose: Aircraft-FGVC \cite{fgvcaircraft} and Stanford Cars \cite{stanfordcars}.

Finally, we experiment on four \textit{industrial anomaly detection} datasets: MVTec-AD~\cite{mvtecad}, MVTec-LOCO~\cite{mvtecloco}, VisA~\cite{visa}, and MPDD~\cite{mpdd}. In these datasets, anomalies are subtle defects, such as scratches, dents, contaminations, and structural changes. MVTec-LOCO also includes logical anomalies, where objects appear in incorrect locations. Unlike the semantic anomaly datasets, anomalies in industrial datasets only cover small sub-regions of the image, with the majority of the image being normal.

\subsection{Training Details}

This section provides an overview of the training details of our proposed pipeline. 

\subsubsection*{Feature Extraction.} In our method, we use the last layer features, followed by normalization using the layer norm component inherent to DINOv2~\cite{oquab2023dinov2}. This normalization step is crucial for stabilizing the feature representation, thereby enhancing the model's ability to process and analyze the input data effectively. We omit image augmentations from our preprocessing steps to ensure our method's general applicability. However, to maintain consistency with the ViT architectures and to optimize input representation, we rescale all input images to a resolution of 518$\times$518.
    
\subsubsection*{Anomalous Feature Generation.} For semantic anomaly detection benchmarks we utilize the distortion strategy \textit{Noise All Patches}. The added noise, \(\epsilon\), follows a Gaussian distribution \(\mathcal{N}(0, 0.25)\). For the industrial anomaly detection benchmarks, the fake features are generated using \textit{Noise Random Patches}. Lastly, for the MVTec-LOCO dataset, which includes logical anomalies, our method uses \textit{Noise Random Patches} and \textit{Attention Shuffle}.
    
\subsubsection*{Discriminator.} To train the discriminator, we use the AdamW optimizer \cite{loshchilov2017decoupled} with an initial learning rate of 0.0005. The learning rate follows a cosine annealing schedule with a decay factor of 0.2. The multi-head attention module in our architecture has 4 heads. The MLP component comprises 3 layers with a hidden dimension of 2048, with no bias added to the last linear layer. The dropout rate after both the attention module and the MLP is set to 0.1. For the semantic anomaly detection datasets, due to their extensive size, we adopted a training regimen of 20 epochs, conducting evaluations after every 250 images. Conversely, for near anomaly detection and industrial anomaly detection datasets, we extended the training duration to 160 epochs, with evaluations after each epoch.
    
\subsubsection*{Inference.} In industrial datasets to detect small defective regions, the distinction between normal and abnormal regions may be minimal, thus we set $K=10$. In contrast, for semantic datasets, we set $K=1369$, which corresponds to all patches for DINOv2, since anomalies typically span the entire image.

\subsection{Main results}

In this section, we showcase our comparative results against state-of-the-art techniques. We specifically target Transformaly~\cite{cohen2022transformaly} and MSAD~\cite{reiss2023mean}, recognized as leading approaches in semantic anomaly detection benchmarks. For near anomaly detection benchmarks, GeneralAD is compared against FITYMI~\cite{mirzaei2023fake}, although we do not use external data or diffusion models to augment the training dataset, unlike FITYMI. Additionally, we include comparisons with SimpleNet~\cite{liu2023simplenet} and Reconstrast~\cite{guo2024recontrast}, which are acknowledged as top-performing methods in industrial anomaly detection. To ensure a fair comparison, we modified the backbone of KDAD~\cite{salehi2021multiresolution} from VGG-16~\cite{simonyan2014very} to a vision transformer and included it as well. We report the AUROC performance for both image-level and pixel-level tasks in tables~\ref{table:sota} and~\ref{table:pixel-sota}. We provide the results of our method with three model sizes of the DINOv2 backbone. Our model performs inference at a rate of 154, 52, and 17 frames per second on an NVIDIA A100-SXM4-40GB GPU for the small, base, and large variants, respectively. 

\subsubsection*{Image-Level Detection.} We present the results in \autoref{table:sota}. As shown, our method not only performed consistently well across various tasks but also surpassed the current state-of-the-art in most of the datasets within different benchmarks. Particularly, we pass Transformaly~\cite{cohen2022transformaly} by $\sim$ 1\% on semantic anomaly detection and by $\sim$6\% on near anomaly detection benchmarks on average. Furthermore, we surpassed FITYMI~\cite{mirzaei2023fake} by $\sim$1\% on near anomaly detection benchmarks. Our method also performed well on industrial anomaly detection datasets, outperforming SimpleNet~\cite{liu2023simplenet} and Recontrast~\cite{guo2024recontrast} on MVTec-LOCO by $\sim$7\% and $\sim$3\% while performing on-par with them on MVTec-AD. 

\begin{table}[H]
  \caption[Image-level AUROC Scores GeneralAD]{\textbf{Image-level AUROC scores.} We compared our method against the current state-of-the-art, encompassing three distinct benchmarks and ten diverse datasets. Unlike other methods that are optimized for peak performance on specific task sets, our approach consistently demonstrated either superior or comparable results across all benchmarks. This underscores the versatility and generality of our proposed method. We report the performance with three model sizes of the DINOv2 backbone. * shows that the method is upgraded with DINOv2-B backbone and reported by us.}
  \label{table:sota}
  \centering
  \scriptsize 
  \resizebox{\linewidth}{!}{%
  \begin{tabular}{@{}lcccccccccc@{}}
    \toprule
    & \multicolumn{4}{c}{Anomaly Detection} & \multicolumn{2}{c}{Near Anomaly Detection} & \multicolumn{4}{c}{Industrial Anomaly Detection} \\
    \cmidrule(lr){2-5} \cmidrule(lr){6-7} \cmidrule(lr){8-11}
    Method & C-10 & C-100 & FMNIST & View & Aircraft & St-Cars & MVAD & MVLOCO & VisA & MPDD \\
    \midrule
    Transformaly & 98.3 & 97.3 & 94.4 & \underline{95.8} & 84.0 & 86.7 & 87.9 & - & - & - \\
    KDAD*         & 98.5 & 97.4 & 94.4 & - & 85.8 & - & 85.8 & 67.5 & 85.7 & 72.1 \\
    MSAD         & 97.2 & 96.4 & 94.2 & - & 79.8 & 87.1 & 85.5 & - & - & - \\
    PANDA        & 96.2 & 94.1 & \textbf{95.6} & 93.6 & 77.7 & \underline{87.6} & 86.5 & - & - & - \\
    FITYMI        & \underline{99.1} & \underline{98.1} & 79.9 & - & 88.7 & \textbf{90.8} & 86.4 & - & - & - \\
    RDAD         & 86.5 & - & 95.0 & - & - & - & 98.4 & 79.7 & \underline{96.0} & 92.7 \\
    Patchcore    & 67.2 & 64.1 & 77.4 & - & 67.8 & 78.3 & 99.2 & 80.3 & 94.2 & 82.1 \\
    SimpleNet    & 86.5 & 69.8 & 87.4 & 76.8 & 83.6 & 81.8 & \textbf{99.6} & 77.6 & 87.9 & 94.8 \\
    Recontrast        & 84.1 & 84.0 & 92.4 & - & - & - & \underline{99.5} & 82.1 & \textbf{97.5} & - \\
    \midrule
    This paper (small)        & 97.7 & 96.6 & 93.9 & \underline{95.8} & 89.5 & 82.2 & 98.7 & 81.9 & 94.4 &  96.2\\
    This paper (base)        & \underline{99.1} & 98.0 & 94.6 & \underline{95.8} & \underline{93.4} & 82.9 & 99.2 & \underline{84.7} & \underline{96.0}  &  \textbf{98.0}\\
    This paper (large)        & \textbf{99.3} & \textbf{98.4} & \underline{95.2} & \textbf{95.9}  & \textbf{94.6} & 87.3 & 99.2 & \textbf{84.9} & 95.9 &  \underline{97.8}\\
    \bottomrule
  \end{tabular}%
}
\end{table}

We attribute this increased performance in the semantic anomaly detection benchmarks compared to previous self-supervised anomaly detection methods like SimpleNet to the differences in the discriminator architecture and training approach. The discriminator takes the features of all the patches of the image as input \mbox{\textit{simultaneously}}, while SimpleNet's discriminator runs inference over every patch \textit{separately}. By including attention among patches, our discriminator detects \textit{global} semantic shifts, which is particularly evident in (near) anomaly detection tasks. Moreover, for industrial anomalies, having access to all the patches in the discriminator allows for adding noise to only a subset of the patches, mimicking the fact that not all the patches in an anomalous image are necessarily anomalous. Furthermore, our model outperforms other methods on logical anomalies due to the self-supervised anomaly feature generation module, which includes adding noise and shuffling a subset of the patches. This shuffling mimics anomalies where objects or components are in the wrong location. 

\subsubsection*{Pixel-Level Localization.}
\begin{figure}[H]
  \centering
  \includegraphics[width=0.9\textwidth]{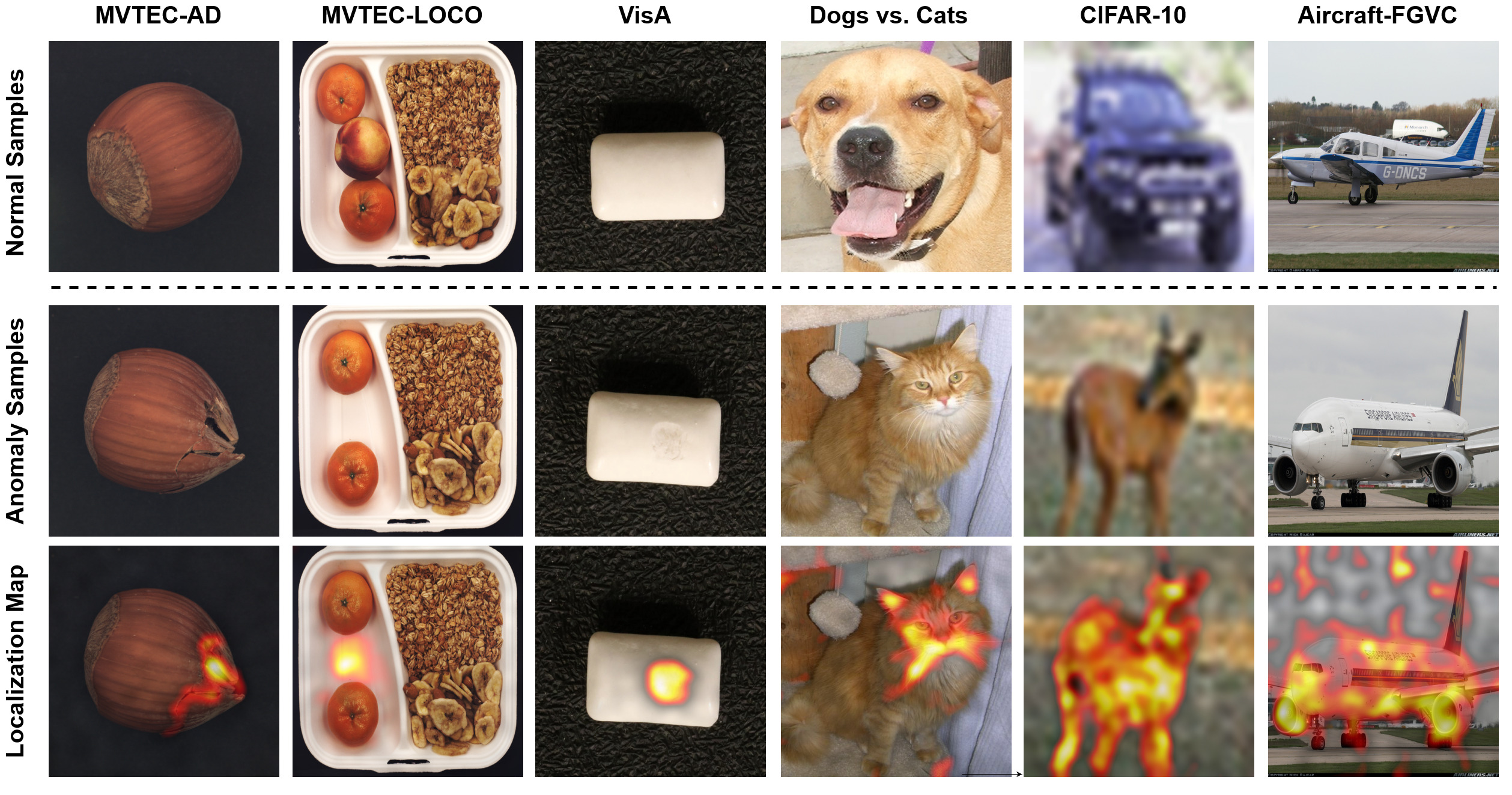}
  \caption{\textbf{Qualitative localization results.} In the first row, normal samples on which the model is trained are shown. In the second row, we show the real anomaly samples; in the third, we show our localization maps. Our method provides interpretable anomaly segmentation maps for both industrial and semantic tasks. For example, when trained on dog images~\cite{elson2007asirra}, it can explain why a cat is an abnormal input. Similarly, when the normal class is cars, the entire object from a different class is localized as an anomaly.}
  \label{fig:localization}
\end{figure}

\begin{figure}[H]
  \centering
  \includegraphics[width=0.9\textwidth]{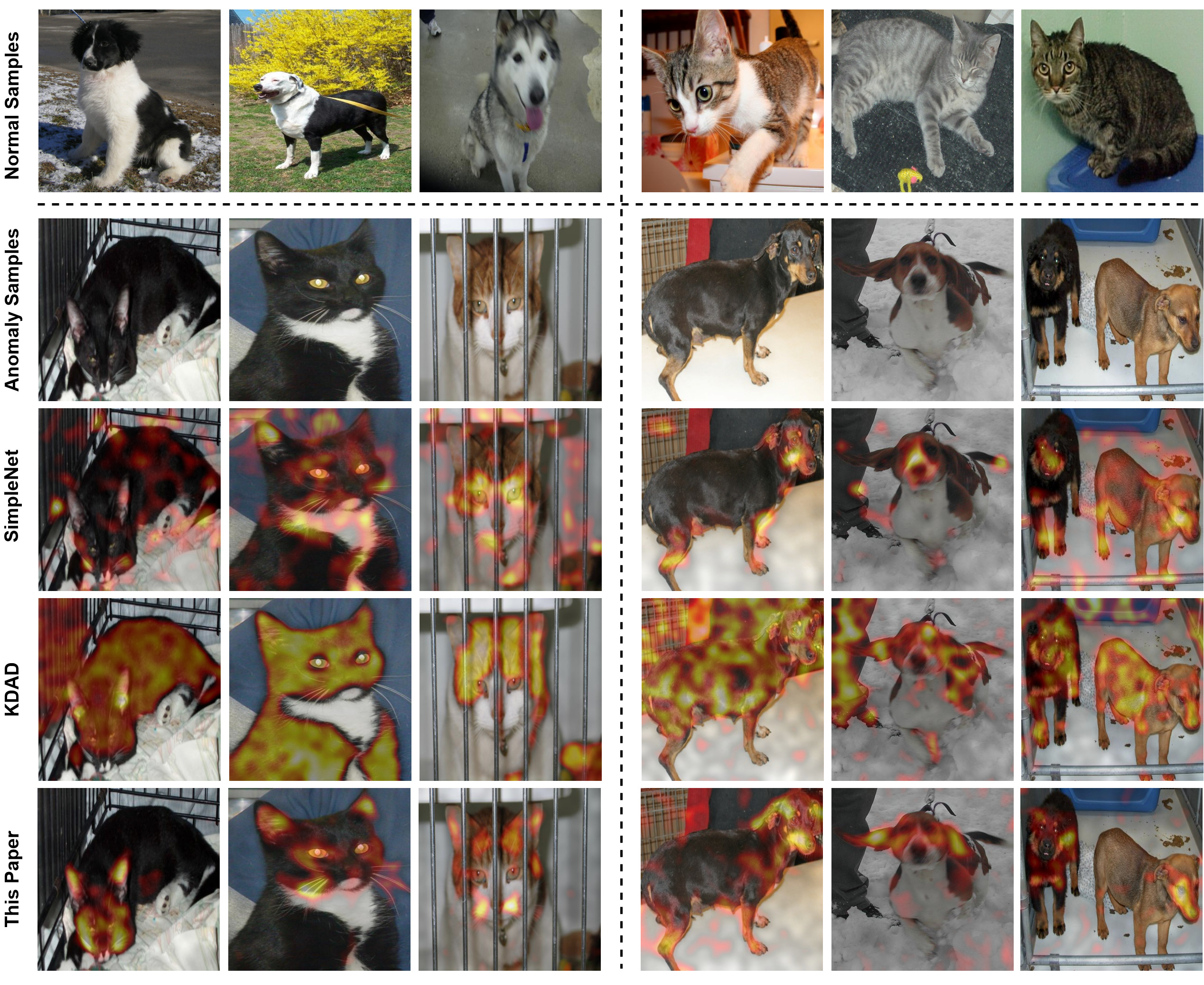}
  \caption{\textbf{Comparison of localization maps.} We conducted a qualitative comparison between the localization maps of our method, SimpleNet, and KDAD. Qualitatively our method shows higher true positive rate and lower false positive rate, thus providing better semantic localization maps.}
  \label{fig:qualitative_comparison}
\end{figure}

We demonstrate the localization capability of our method in \autoref{table:pixel-sota}. While almost all state-of-the-art methods in semantic benchmarks perform poorly on anomaly localization, the proposed method shows strong performance in this task. GeneralAD achieves state-of-the-art pixel-level performance on VisA and performs on par with other methods on MVTec-AD, despite not being specifically designed for such datasets. Furthermore, our localization is not restricted to industrial tasks and can be employed to improve the interpretability of decisions for semantic datasets, as shown in \autoref{fig:localization} and \ref{fig:qualitative_comparison}.

\begin{table}[H]
\caption[Pixel-level AUROC Scores GeneralAD]{\textbf{Pixel-level AUROC scores.} Our method performed close to state-of-the-art methods that are specifically designed for industrial benchmarks. As opposed to such methods, we were able to provide interpretable maps for semantic tasks as well, demonstrating that the method is more general and applicable across different tasks.}
\label{table:pixel-sota}
\centering
\begin{tabular}{lccccccc}
\toprule
Method & RDAD & Patchcore & SimpleNet & Recontrast & \multicolumn{3}{c}{This paper} \\
 \cmidrule(lr){6-8} 
& &&& & \hspace{1pt}  small \hspace{1pt} & \hspace{1pt} base \hspace{1pt} & \hspace{1pt} large \hspace{1pt}\\
\midrule
MVTec-AD & 97.8 & \underline{98.1} & \underline{98.1} & \textbf{98.4} & 97.0 & 97.2 & 97.7\\
VisA & 90.1 & \underline{98.8} & 91.8 & 98.2 & 98.2 & \underline{98.8} & \textbf{99.0}\\
\midrule
Mean & 94.0 & \textbf{98.5} & 95.0 & 98.3 & 97.6 & 98.0 & \underline{98.4}\\
\bottomrule
\end{tabular}
\end{table}

In \autoref{fig:localization}, we show the qualitative results of our method across a wide range of anomaly detection tasks. As shown, the method provides interpretable maps for semantic datasets, which can improve safety and reliability when deployed in real-world scenarios, such as self-driving cars. Furthermore, it produces precise anomaly segmentation maps for logical and structural anomalies when evaluated on industrial tasks. This supports the generality of our method, which can be used in a wide range of applications. 

In \autoref{fig:qualitative_comparison}, we qualitatively compare GeneralAD against state-of-the-art methods KDAD~\cite{salehi2021multiresolution} and SimpleNet~\cite{liu2023simplenet} on semantic anomaly localization. KDAD, despite outperforming other methods in image-level semantic anomaly detection, generates a high number of false positive pixels in the localization map. 
SimpleNet, on the other hand, does not consistently focus on the most relevant parts, resulting in lower true positive results, albeit with lower false positive rates. Our method, however, infers the semantic correlation between different patches and produces localization maps with a high true positive rate and low false positive rate, demonstrating its superiority in anomaly localization.

\subsubsection*{Few-shot Anomaly Detection.} Few-shot anomaly detection (FSAD) has been introduced to address the needs of quick manufacturing transitions~\cite{salehi2021unified}. Despite involving a training stage, our pipeline consistently outperforms methods like PatchCore~\cite{roth2022towards} by large margins.  This demonstrates that GeneralAD is also sample-efficient, making it more practical for real-world applications.

\begin{table}[H]
    \centering
    \caption{\textbf{Few-shot anomaly detection AUROC scores on MVTec AD.} Each column indicates the number of training samples used in the experiment. The results are reported over 5 random seeds.}
    \setlength{\tabcolsep}{3pt}
    \begin{tabular}{lcccc}
        \toprule
         Shots & 1 & 2 & 4 & 8 \\
        \midrule
        SPADE & 71.6 & 73.4 & 82.8 & 84.0\\
        PaDiM & 76.1 & 78.9 & 80.5 & 82.0\\
        PatchCore & 84.1 & 87.2 & 88.5 & 92.2\\
        \midrule
        This paper (small)& 84.4{\scriptsize $\pm$1.0} & 86.8{\scriptsize $\pm$0.4} & 90.0{\scriptsize $\pm$1.2} & 90.5{\scriptsize $\pm$1.4} \\
        This paper (base)& \underline{87.0{\scriptsize $\pm$1.4}} & \textbf{91.9{\scriptsize $\pm$0.9}} & \textbf{93.1{\scriptsize $\pm$0.8}} & \underline{93.5{\scriptsize $\pm$0.8}}\\
        This paper (large)& \textbf{87.5{\scriptsize $\pm$2.2}} & \underline{91.5{\scriptsize $\pm$1.4}}& \underline{92.8{\scriptsize $\pm$1.7}}& \textbf{93.6{\scriptsize $\pm$0.6}}\\
        \bottomrule
    \end{tabular}
    \label{tab:few_shot}
\end{table}

\subsection{Ablations} 

\subsubsection*{Pretraining and Backbone.} To demonstrate that our results are not solely due to the use of a state-of-the-art backbone but rather the combination of the backbone and our method, we evaluated our method and other state-of-the-art methods using different backbone architectures \cite{dosovitskiy2020image,radford2021learning,cherti2023reproducible,caron2021emerging,darcet2023vision,zagoruyko2016wide}. The results are presented in \autoref{tab:backbone}. Semantic state-of-the-art methods such as KDAD and Transformaly significantly fall behind the state-of-the-art on industrial datasets regardless of the backbone. Similarly, SimpleNet with different backbones underperforms in the semantic benchmarks. This indicates that such methods have been designed to use specific aspects of the pretrained features by optimizing for particular benchmarks. However, our method was designed to leverage the semantic features of DINOv2~\cite{oquab2023dinov2} specifically and thus perform consistently across different datasets.  

\begin{table}[H]
\caption[Ablation Study GeneralAD: Backbone]{\textbf{Effect of pretraining and backbone.} We replaced the backbone of different state-of-the-art methods with DINOv2-B. As shown, no other method could excel in all the benchmarks by varying pretraining. GeneralAD, instead, could effectively exploit the features of DINOv2 and worked generally well across different tasks.}
\label{tab:backbone}
\centering
\scriptsize
\begin{tabular}{@{}llcccc@{}}
\toprule
Method & Backbone & CIFAR-10 & Aircraft-FGVC & MVTec-AD & MVTec-LOCO \\
\midrule
Transformaly & ViT-B/16 & 98.3 & 84.0 & 87.9 & 65.4 \\
 & DINOv2-B/14 & 98.0 & 82.3 & 80.9 & 65.0 \\
KDAD & ViT-B/16 & 98.1 & 86.6 & 80.2 & 67.5 \\
 & DINOv2-B & 98.5 & 85.8 & 85.8 & 60.5 \\
SimpleNet & WideResNet50 & 86.5 & 83.6 & \textbf{99.6} & 77.6 \\
 & ViT-B/16 & 85.8 & 83.7 & 93.3 & 78.4 \\
 & DINOv2-B/14 & 83.7 & \underline{88.7} & 97.7 & 81.7 \\
This paper & ViT-B/16 & 93.0 & 85.4 & 83.8 & 68.1 \\
& OpenClip-B/14 & 93.9 & 83.7 & 98.5 &  81.9\\
& DINO-B/8 & 89.5 & 75.2 & 98.2 & 79.7 \\
& DINOv2-B-reg4/14 & \textbf{99.3} & \textbf{93.4} & 98.8 & \underline{83.2} \\
& DINOv2-B/14 & \underline{99.1} & \textbf{93.4} & \underline{99.2} & \textbf{84.7} \\
\bottomrule
\end{tabular}
\end{table}

\subsubsection*{$K$ and Noise Magnitude.} 
\begin{figure}[H]
  \centering
  \includegraphics[width=0.95\textwidth]{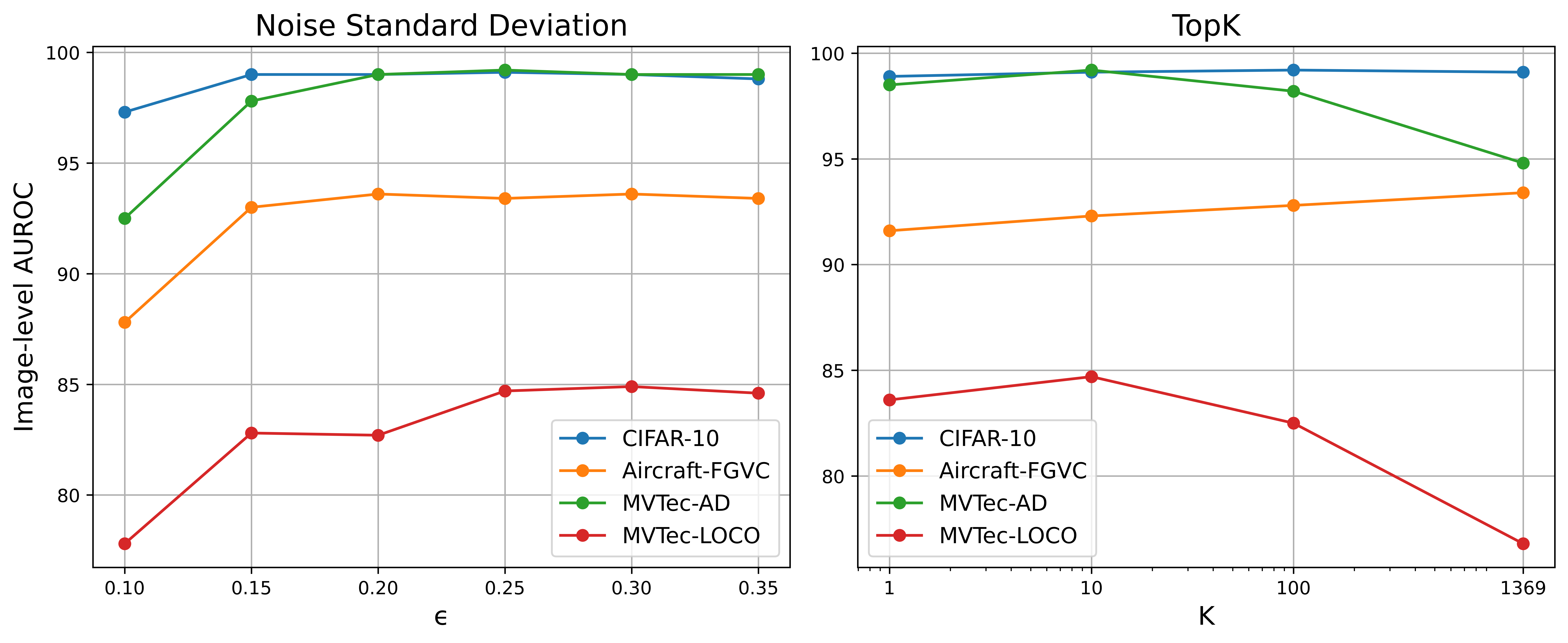}
  \caption{\textbf{The effect of $K$ and noise magnitude.} Independent of the type of anomaly, the best performance is found with a moderate amount of Gaussian noise. Therefore, we select \(\epsilon = 0.25\) for all experiments. The optimal top \(K\) parameter depends on the size of the anomalies in the dataset; thus, we choose \(K=1369\) for semantic (near) anomaly detection and \(K=10\) for industrial anomaly detection.}
  \label{fig:hyperparameters}
\end{figure}

We evaluate the effect of \(K\) in the top \(K\) selection and the Gaussian noise magnitude (\(\epsilon\)) in \autoref{fig:hyperparameters}. The results indicate that the method performs consistently across a range of Gaussian noise magnitudes, with the optimal magnitude of 0.25 across different anomaly distributions. For the top \(K\) parameter, it is essential to set \(K\) close to the size of the anomalies in the input. As shown in \autoref{fig:hyperparameters}, for (near) anomaly detection, where anomalies span almost the entire image, this leads to \(K=1369\). For industrial anomaly detection, characterized by subtle defects in small parts of the image, this results in \(K=10\).

\subsubsection*{Distortion Types.} In \autoref{tab:abl_distortion_type}, we show the effect of different feature distortion types across various benchmarks. As shown, each dataset distribution can benefit from specific types of distortions. For semantic datasets, adding noise to all patches gives the best performance, likely due to the distribution of objects where anomalies can cover the entire image. For datasets with anomalies in small regions, such as MVTec-AD, adding noise to random locations performs best. This helps the discriminator identify anomalies within a complex industrial context where normal and abnormal patches coexist.

For datasets with logical anomalies, such as MVTec-LOCO, using more complex distortions, such as shuffling, is beneficial. This technique targets patches with high attention scores, representing the importance of each patch feature, and shuffles them to create anomalies. The model then learns to recognize logical anomalies based on disrupted spatial relations.

\begin{table}[H]
    \centering
    \caption[Ablation Study GeneralAD: Distortion Type]{\textbf{The effect of distortion types.} Different distortion strategies are optimal for different types of anomalies. For semantic datasets, the most effective approach is \textit{Noise All Patches}. In contrast, for industrial datasets that primarily contain structural defects, such as MVTec-AD, \textit{Noise Random Patches} is the most suitable. Finally, for datasets containing logical anomalies, incorporating \textit{Attn Shuffle} proves to be the most effective.}
    \begin{tabular}{lccccc}
        \toprule
        Distortion & CIFAR-10 & Aircraft-FGVC & MVTec-AD & MVTec-LOCO  \\
        \midrule
        Noise All Patches & \textbf{99.3} & \textbf{94.6} & \underline{99.0} & \underline{84.1} \\
        Noise Random Patches  & \underline{96.6} & \underline{94.3} & \textbf{99.2} & 83.8 \\
        \hspace{5pt} + Attention Shuffle & 93.6 & 74.5 & \underline{99.0} & \textbf{84.9} \\
        \bottomrule
    \end{tabular}
    \label{tab:abl_distortion_type}
\end{table}

\section{Conclusion}
\label{sec:conclusion}
In this paper, we have proposed a new approach that significantly narrows the methodological chasm between semantic, near-distribution and industrial benchmarks.
Our method leverages a pretrained Vision Transformer (DINOv2) feature extractor alongside a novel self-supervised anomaly feature generation module. This methodology facilitates the creation of pseudo-abnormal samples with subtle, challenging distortions and employs a transformer-based discriminator capable of detecting a wide range of anomalies. It achieves state-of-the-art results in six out of the ten datasets. For the remaining four datasets, our method performs on par with existing standards. This is accomplished without the need for extensive per-task adjustments, in stark contrast to existing works. Finally, our method facilitates the generation of interpretable localization maps, enhancing the understanding and analysis of detected anomalies. 
\textit{Limitations.} Despite our effort to introduce a generic method that works across all domains, our method does not yet succeed wholly across the board. We believe that the set of diverse benchmarks that we have evaluated in this paper can serve as a springboard for a generation of new and general anomaly detection methods.

\section*{Acknowledgements}
This research was funded by the University of Amsterdam. The authors acknowledge SURF for providing access to the National Supercomputer Snellius.

\bibliographystyle{splncs04}
\bibliography{main}

\setcounter{section}{0}
\renewcommand{\thesection}{\Alph{section}}

\section{Datasets}
\label{appendix:datasets}
\label{app_datasets}

\begin{figure}
  \centering
  \includegraphics[width=0.99\linewidth]{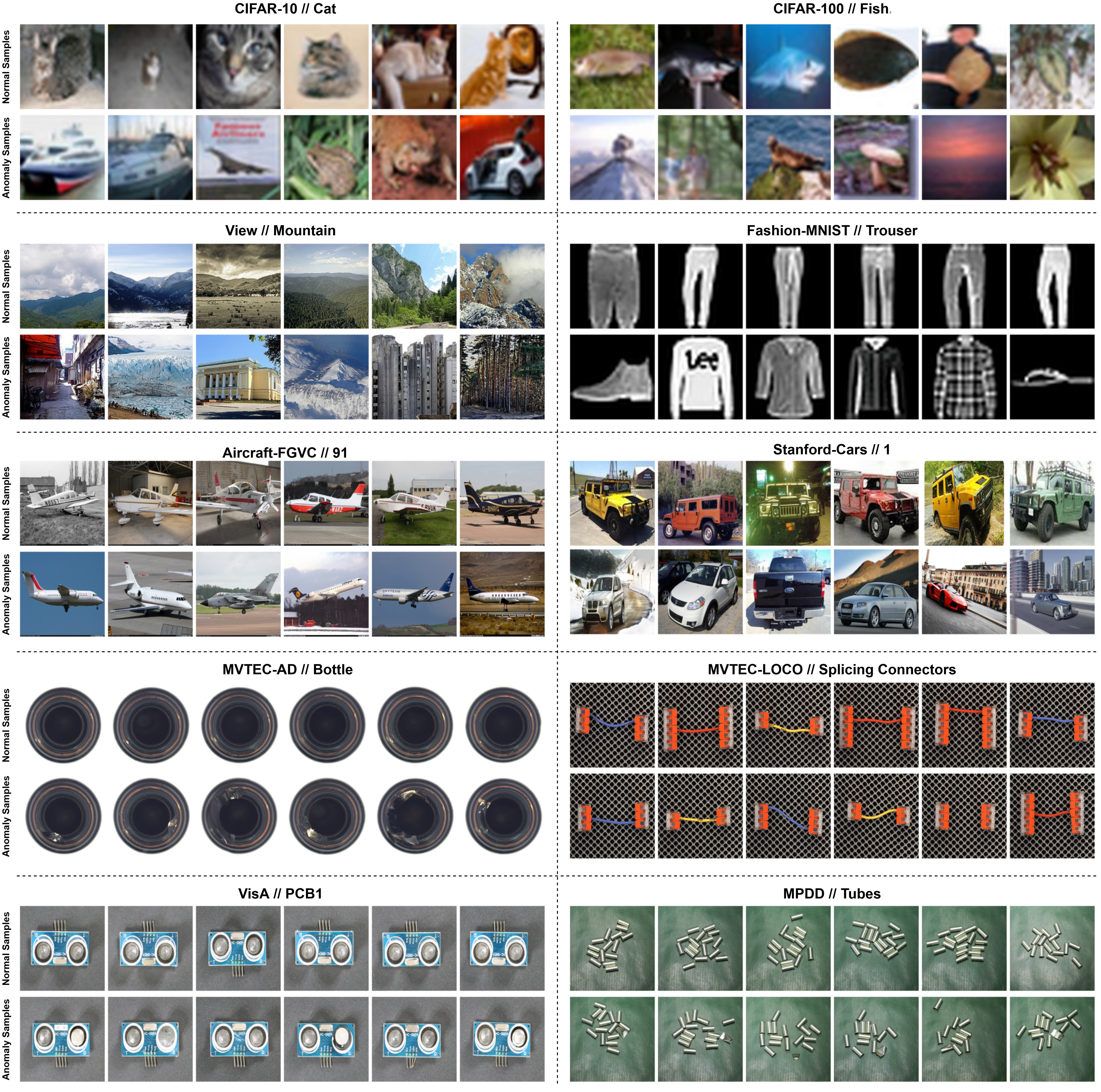}
  \caption{\textbf{Normal and Anomaly samples from all datasets.}}
  \label{fig:dataset}
\end{figure}

Here we show a set of normal and anomaly samples that can be used in our experiments. As Figure~\ref{fig:dataset} shows, we have conducted diverse and comprehensive experiments across different datasets to support the generality of our approach. Further details on the datasets are given below.

\subsubsection*{CIFAR~\cite{cifar}:} CIFAR-10 and CIFAR-100 consist of 60,000 natural color images each, with a resolution of 32×32. These images are divided into a training set of 50,000 and a testing set of 10,000. CIFAR-10 is structured into 10 equally sized classes, while CIFAR-100 is split into either 100 fine-grained or 20 coarse-grained classes, with our experiments following the coarse-grained classification.

\subsubsection*{Fashion MNIST~\cite{fmnist}:} FMNIST comprises 60,000 training samples and 10,000 test samples, each being a 28×28 grayscale image distributed among 10 unique classes.

\subsubsection*{View~\cite{view}:} The View dataset is a collection of natural scene images divided into six classes, such as mountains and buildings. This dataset provides approximately $\sim$2,300 images per class for training and $\sim$500 images per class for testing. 

\subsubsection*{Dogs vs. Cats~\cite{elson2007asirra}:} This simple visualization dataset contains 25,000 images of dogs and cats. 

\subsubsection*{Aircraft-FGVC~\cite{fgvcaircraft}:} This dataset comprises 10,200 images, each representing one of 102 different aircraft model variants, predominantly airplanes, with each variant having 100 images. An 80-20 train-test split is used. We evaluated the following ten classes: [91,96,59,19,37,45,90,68,74,89], aligning our approach with \cite{mirzaei2023fake}.

\subsubsection*{Stanford-Cars~\cite{stanfordcars}} This dataset includes 16,185 images spread across 196 car classes. The dataset is divided into 8,144 training images and 8,041 testing images, with an approximately equal split for each class. We ran our experiments on the first 20 classes and tested on the entire test set, aligning our approach with \cite{mirzaei2023fake}.

\subsubsection*{MVTec-AD~\cite{mvtecad}:} The MVTec Anomaly Detection dataset is recognized for its collection of 5354 high-resolution images from 15 different categories of industrial objects and textures, serving as the main testbed for anomaly detection algorithms in manufacturing. The defects present in these anomalies are varied, encompassing more than 70 different kinds, including scratches, dents, contaminations, and structural changes. 

\subsubsection*{MVTec-LOCO~\cite{mvtecloco}:} This dataset targets anomaly localization with images of industrial products that contain logical and structural defects. It contains 3644 images from five different classes inspired by industrial inspection scenarios. 

\subsubsection*{VisA~\cite{visa}:} The Visual Anomaly dataset is the largest industrial anomaly benchmark with images of various manufacturing anomalies. It contains 12 classes in 3 domains across 10,821 high-resolution images.

\subsubsection*{MPDD~\cite{mpdd}:} This smaller dataset is specifically designed to address the challenges of detecting defects in the fabrication of painted metal parts. It offers a realistic testing environment with variable spatial orientations, multiple objects, and non-homogeneous backgrounds, diverging from traditional lab-based AD datasets. It contains 1346 images across 6 categories.

\section{Extended Qualitative Results}
\label{appendix:localization_maps}
\begin{figure}
  \centering
  \includegraphics[width=0.99\textwidth]{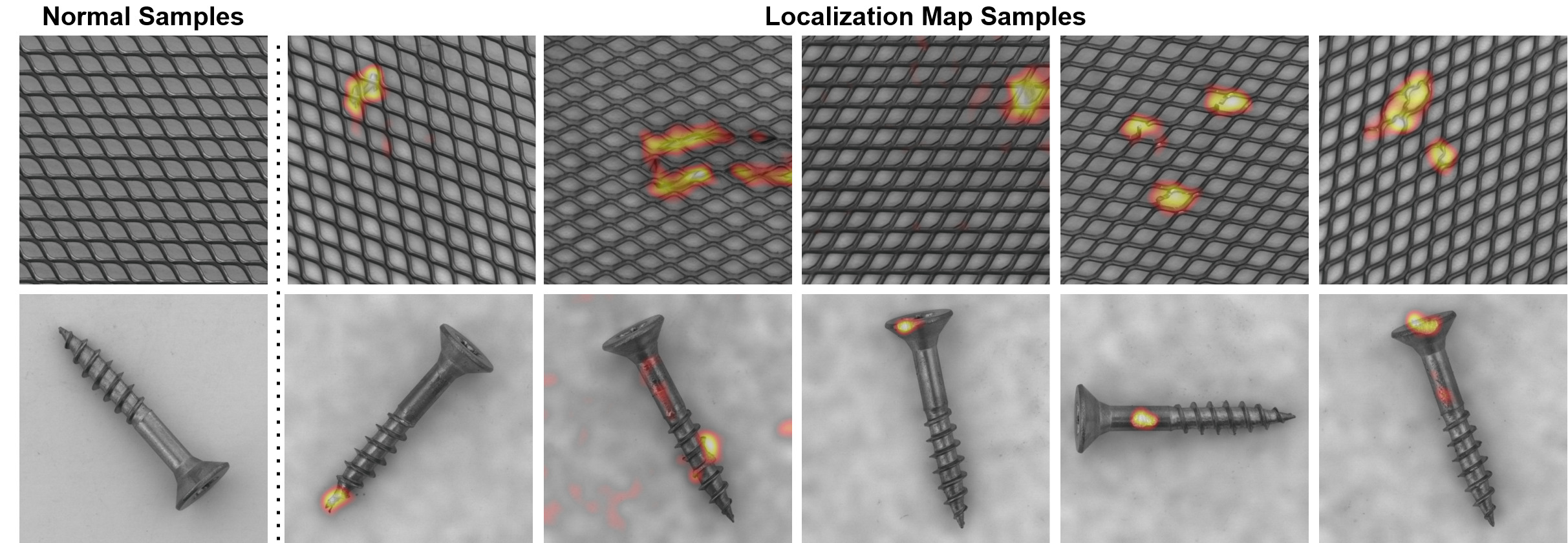}
  \caption[Appendix GeneralAD: Extended Industrial Qualitative Results]{\textbf{Extended qualitative localization results GeneralAD on MVTec-AD.} We train the method on the normal samples specified in the first column, then specify the abnormality regions for each input pass at the test time.}
  \label{fig:mvtec_localization}
\end{figure}

\begin{figure}
  \centering
  \includegraphics[width=0.99\textwidth]{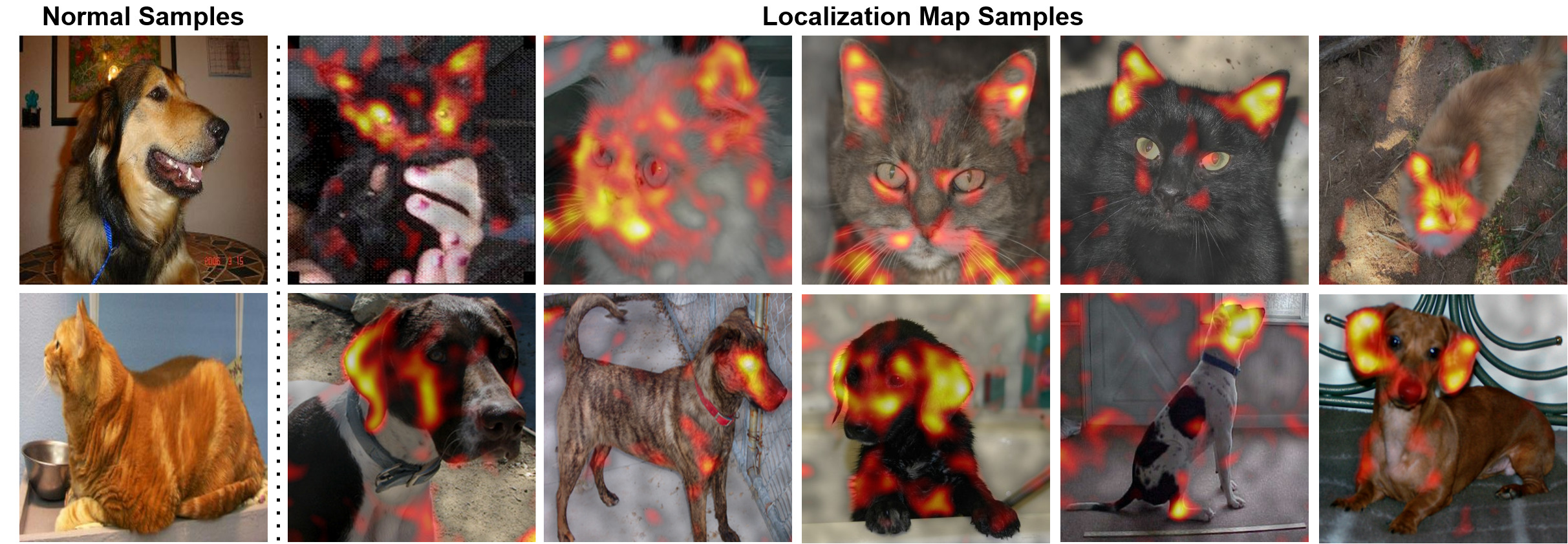}
  \caption[Appendix GeneralAD: Extended Semantic Qualitative Results]{\textbf{Extended qualitative localization results GeneralAD on Dogs vs. Cats.} We train the method on the normal samples specified in the first column, then specify the abnormality regions for each input pass at the test time. As is shown, the method focuses mainly on the snout, whiskers, and ears to discriminate between dogs and cats. }
  \label{fig:localization_1}
\end{figure}

We show more qualitative results of our method. \autoref{fig:mvtec_localization} shows the localization results of our method on two classes of MVTec AD~\cite{mvtecad}. Our method can accurately detect subtle defects in the dataset, such as in screw images. \autoref{fig:localization_1} shows the localization results of our method on the Dogs vs. Cats dataset \cite{elson2007asirra}. As can be seen, our method consistently focuses on discriminative features like the snout, whiskers, and ears.

\section{Per Class Results}
\label{appendix:classes}
Here we show the per-class results of our main results table. Table~\ref{table:per_class_sota_semantic} shows the results for the semantic anomaly detection datasets, Table~\ref{table:per_class_sota_near} for near anomaly detection datasets, and Table~\ref{table:per_class_sota_industrial} for industrial anomaly detection datasets.

\begin{table}
  \caption{\textbf{Image-level AUROC scores for GeneralAD on the semantic anomaly detection datasets.} The per-class results of GeneralAD on the datasets CIFAR-10 and CIFAR-100~\cite{cifar}, FMNIST~\cite{fmnist}, and View~\cite{view}.}
  \label{table:per_class_sota_semantic}
  \centering
  \scriptsize 
  \resizebox{\linewidth}{!}{
  \begin{tabular}{@{}lcccccccccc@{}}
    \toprule
     \multicolumn{2}{c}{CIFAR-10} & \multicolumn{2}{c}{CIFAR-100} & \multicolumn{2}{c}{FMNIST} & \multicolumn{2}{c}{View}\\
    Class & I-AUROC (\%) & Class & I-AUROC (\%) & Class & I-AUROC (\%) & Class & I-AUROC (\%) \\
    \midrule
    airplane & 99.9 & aquatic mammals & 97.4 & t-shirt/top & 92.2 & buildings & 93.2 \\
    automobile & 99.5 & fish & 98.1 & trouser & 99.0 & forest & 99.6 \\
    bird & 99.5 & flowers & 99.7 & pullover & 93.1 & glacier & 94.5 \\
    cat & 97.7 & food containers & 99.3 & dress & 95.2 & mountain & 94.5 \\
    deer & 99.1 & fruit and vegetables & 99.5 & coat & 93.3 & sea & 95.7 \\
    dog & 98.5 & household electrical devices & 98.2 & sandal & 97.9 & street & 98.1 \\
    frog & 99.8 & household furniture & 99.5 & shirt & 85.4 &  &  \\
    horse & 99.5 & insects & 99.1 & sneaker & 98.3 &  &  \\
    ship & 99.8 & large carnivores & 97.9 & bag & 99.1 &  &  \\
    truck & 99.8 & large man-made outdoor things & 96.7 & ankle boot & 98.4 &  &  \\
    &  & large natural outdoor scenes & 98.0 &  &  &  &  \\
    &  & large omnivores and herbivores & 97.5 &  &  &  &  \\
    &  & medium-sized mammals & 96.2 &  &  &  &  \\
    &  & non-insect invertebrates & 98.6 &  &  &  &  \\
    &  & people & 99.4 &  &  &  &  \\
    &  & reptiles & 98.6 &  &  &  &  \\
    &  & small mammals & 98.1 &  &  &  &  \\
    &  & trees & 98.4 &  &  &  &  \\
    &  & vehicles 1 & 98.7 &  &  &  &  \\
    &  & vehicles 2 & 98.5 &  &  &  &  \\
    \midrule
    All avg.  & 99.3 &  & 98.4 &  & 95.2 &  & 95.9 \\
    \bottomrule
  \end{tabular}
  }
\end{table}

\begin{table}
  \caption{\textbf{Image-level AUROC scores for GeneralAD on the near anomaly detection datasets.} The per-class results of GeneralAD on the datasets Aircraft-FGVC~\cite{fgvcaircraft} and Stanford-Cars~\cite{stanfordcars}.}
  \label{table:per_class_sota_near}
  \centering
  \begin{tabular}{@{}lcccccccccc@{}}
    \toprule
     \multicolumn{2}{c}{Aircraft-FGVC} & \multicolumn{2}{c}{Stanford-Cars}\\
    Class & I-AUROC (\%) & Class & I-AUROC (\%) \\
    \midrule
    91 & 98.7 & 1 & 98.5\\
    96 & 99.5 & 2 & 76.6\\
    59 & 90.5 & 3 & 90.4\\
    19 & 91.8 & 4 & 86.9\\
    37 & 99.4 & 5 & 87.9\\
    45 & 91.8 & 6 & 83.9\\
    90 & 96.4 & 7 & 76.2\\
    68 & 90.7 & 8 & 82.7\\
    74 & 92.8 & 9 & 78.6\\
    89 & 94.5 & 10 & 91.5\\
     &  & 11 & 91.1\\
     &  & 12 & 86.1\\
     &  & 13 & 91.4\\
     &  & 14 & 88.5\\
     &  & 15 & 87.5\\
     &  & 16 & 97.5\\
     &  & 17 & 90.8\\
     &  & 18 & 89.9\\
     &  & 19 & 84.8\\
     &  & 20 & 86.0\\
    \midrule
    All avg.  & 94.6 &  & 87.3\\
    \bottomrule
  \end{tabular}
\end{table}

\begin{table}
  \caption{\textbf{Image-level AUROC scores for GeneralAD on the industrial anomaly detection datasets.} The per-class results of GeneralAD on the datasets MVTec-AD~\cite{mvtecad}, MVTec-LOCO~\cite{mvtecloco}, VisA~\cite{visa}, and MPDD~\cite{mpdd}.}
  \label{table:per_class_sota_industrial}
  \centering
  \scriptsize 
  \resizebox{\linewidth}{!}{%
  \begin{tabular}{@{}lcccccccccc@{}}
    \toprule
     \multicolumn{2}{c}{MVTec-AD} & \multicolumn{2}{c}{MVTec-LOCO} & \multicolumn{2}{c}{VisA} & \multicolumn{2}{c}{MPDD}\\
    Class & I-AUROC (\%) & Class & I-AUROC (\%) & Class & I-AUROC (\%) & Class & I-AUROC (\%) \\
    \midrule
    tile & 100 & screw bag & 74.4 & candle & 95.3 & bracket black & 93.8 \\
    bottle & 100 & pushpins & 77.1 & capsules & 93.4 & bracket brown & 97.2 \\
    cable & 98.8 & juice bottle & 93.7 & cashew & 92.9 & bracket white & 99.4 \\
    capsule & 97.6 & breakfast box & 90.9 & chewinggum & 99.4 & connector &  96.7\\
    carpet & 99.8 & splicing connectors & 88.2 & fryum & 95.7 & metal plate & 100 \\
    grid & 100 & & & macaroni1 & 96.8 & tubes & 99.7 \\
    hazelnut & 99.9 & & & macaroni2 & 89.1 & & \\
    leather & 100 & & & pcb1 & 97.2 & & \\
    metal nut & 100 & & & pcb2 & 97.7 & & \\
    pill & 95.1 & & & pcb3 & 96.5 & & \\
    screw & 96.9 & & & pcb4 & 99.3 & & \\
    toothbrush & 100 & & & pipe fryum & 97.7 & & \\
    transistor & 100 & & & & & & \\
    wood & 99.9 & & & & & & \\
    zipper & 100 & & & & & & \\
    \midrule
    All avg.  & 99.2 &  & 84.9 &  & 95.9 &  & 97.8 \\
    \bottomrule
  \end{tabular}
  }
\end{table}

\end{document}